%% file: main.tex
\documentclass[runningheads]{llncs}

\usepackage{graphicx}
\usepackage{subcaption}
\usepackage{comment}
\usepackage{makecell}
\usepackage{amsmath,amssymb}
\usepackage{bbm}
\usepackage{color}
\usepackage[USenglish]{babel}
\usepackage{booktabs,multirow}
\usepackage{tablefootnote}
\usepackage{placeins}

\usepackage{makeidx}

\usepackage{hyperref}

\input{commands}
\begin{document}
\pagestyle{headings}
\mainmatter

\title{Learning Monocular 3D Vehicle Detection without 3D Bounding Box Labels}
\titlerunning{Learning 3D Vehicle Detection without 3D Bounding Box Labels}
\authorrunning{Lukas Koestler, Nan Yang, Rui Wang, and Daniel Cremers}
\author{%
Lukas Koestler\textsuperscript{1,2}\thanks{%
\url{lukas.koestler@tum.de}, project page: \url{https://lukaskoestler.com/ldwl}%
}\hspace{8pt}%
Nan Yang\textsuperscript{1,2}\hspace{8pt}%
Rui Wang\textsuperscript{1,2}\hspace{8pt}%
Daniel Cremers\textsuperscript{1,2}%
}
\institute{\textsuperscript{1}Technical University of Munich\hspace{16pt}\textsuperscript{2}Artisense}

\maketitle

\begin{abstract}
The training of deep-learning-based 3D object detectors requires large datasets with 3D bounding box labels for supervision that have to be generated by hand-labeling. We propose a network architecture and training procedure for learning monocular 3D object detection without 3D bounding box labels. By representing the objects as triangular meshes and employing differentiable shape rendering, we define loss functions based on depth maps, segmentation masks, and ego- and object-motion, which are generated by pre-trained, off-the-shelf networks. We evaluate the proposed algorithm on the real-world KITTI dataset and achieve promising performance in comparison to state-of-the-art methods requiring 3D bounding box labels for training and superior performance to conventional baseline methods.
\end{abstract}

\section{Introduction}\label{sec:introduction}
\sloppy Three-dimensional object detection is a crucial component of many autonomous systems because it enables the planning of collision-free trajectories. Deep-learning-based approaches have recently shown remarkable performance \cite{wang_pseudo-lidar_2019} but require large datasets for training. More specifically, the detector is supervised with 3D bounding box labels which are obtained by hand-labeling LiDAR point clouds \cite{geiger_are_2012}. On the other hand, methods that optimize pose and shape of individual objects utilizing hand-crafted energy functions do not require 3D bounding box labels \cite{engelmann_joint_2016,wang2020directshape}. However, these methods cannot benefit from training data and produce worse predictions in our experiments. To leverage deep learning and overcome the need for hand-labeling, we thus introduce a training scheme for monocular 3D object detection which does not require 3D bounding box labels for training.

We build upon \pseudolidar~\cite{wang_pseudo-lidar_2019}, a recent supervised 3D object detector that utilizes a pre-trained image-to-depth network to back-project the image into a point cloud and then applies a 3D neural network. To replace the direct supervision by 3D bounding box labels, our method additionally uses 2D instance segmentation masks, as well as, ego- and object-motion as inputs during training. We show that our method works with off-the-shelf, pre-trained networks: \maskrcnn~\cite{he_mask_2017} for segmentation and \structtodepth~\cite{casser_depth_2019} for motion estimation. Therefore, we introduce no additional labeling requirements for training in comparison to \pseudolidar. During inference the motion network is not required.

\begin{figure}[tb]
\centering
\includegraphics[width=\textwidth]{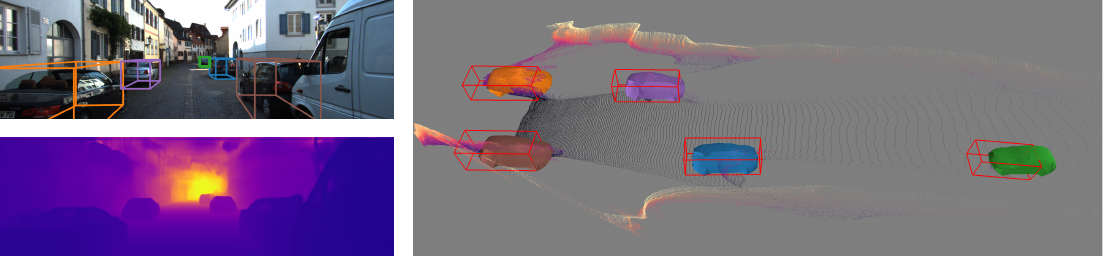}
\caption[Detection visualization]{We propose a monocular 3D vehicle detector that requires no 3D bounding box labels for training. The right image shows that the predicted vehicles (\textit{colored shapes}) fit the ground truth bounding boxes (\textit{red}). Despite the noisy input depth (\textit{lower left}), our method is able to accurately predict the 3D poses of vehicles due to the proposed fully differentiable training scheme. We additionally show the projections of the predicted bounding boxes (\textit{colored boxes, upper left}).}
\label{fig:teaser}
\end{figure}

\sloppy Due to the \pseudolidar-based architecture, our approach can utilize depth maps from mono-to-depth, or stereo-to-depth methods, which can be self-supervised or supervised. We show experiments for all four combinations. For depth maps generated by a self-supervised mono-to-depth network~\cite{godard_digging_2019}, only \maskrcnn{} needs to be trained supervisedly and we use a model pre-trained on the general COCO dataset~\cite{coco}, therefore avoiding any supervision on the KITTI dataset.

\subsection{Related Work}
\subsubsection{Object Detection.}
Two-dimensional object detection is a fundamental task in computer vision, where two-stage, CNN-based detectors~\cite{ren_faster_2015} have shown impressive performance. \maskrcnn~\cite{he_mask_2017} extends this approach to include the prediction of instance segmentation masks with high accuracy.

In contrast, image-based 3D object detection is still an open problem because depth information has to be inferred from 2D image data.
Approaches based on per-instance optimization minimize a hand-crafted energy function for each object individually; the function encodes prior knowledge about pose and shape and considers input data, e.g., the back-projection of an estimated depth map~\cite{engelmann_joint_2016}, an image-gradient-based fitness measure \cite{zhang_three-dimensional_2011}, or the photometric constraint for stereo images together with 2D segmentation masks~\cite{wang2020directshape}.
Initial deep-learning-based methods for stereo images~\cite{chen_3d_2015} and monocular images~\cite{chen_monocular_2016} generate object proposals which are then ranked by a neural network. Subsequent approaches employ geometric constraints to lift 2D detections into 3D~\cite{mousavian_3d_2017,qin_monogrnet_2019}. Kundu et al.~\cite{kundu_3d-rcnn_2018} propose to compare the predicted pose and shape of each object to the ground truth depth map and segmentation mask, which yields two additional loss terms during training. They employ rendering to define the loss function and approximate the gradient using finite differences. Their approach relies on 3D bounding box labels for supervision and uses the additional loss terms to improve the final performance.
Li et al.~\cite{li_stereo_2019} propose \stereorcnn{} which combines deep learning and per-instance optimization for object detection from stereo images. Similar to our approach, \stereorcnn{} does not supervise the 3D position using 3D bounding box labels. In contrast to our method, they use the 3D bounding box labels to directly supervise the 3D dimensions, the viewpoint, and the perspective keypoint. Replacing the 3D bounding box labels by estimated 3D dimensions, viewpoints, and perspective keypoints is a non-trivial extension of their work. Furthermore, it is not studied how well their algorithm would handle the inevitable noise in the estimated 3D dimensions, viewpoints, and perspective keypoints if they are not computed from the highly accurate ground truth labels. Moreover, \stereorcnn{} is designed specifically for stereo images, while the proposed method is designed for monocular images and can be easily extended to the stereo setting (cf. \autoref{sec:experiments}).
Wang et al.~\cite{wang_pseudo-lidar_2019} back-project the depth map obtained from an image-to-depth network to a point cloud and then use networks initially designed for LiDAR data~\cite{qi_frustum_2018,ku_joint_2018} for detection. Their method, \pseudolidar, showed that representing depth information in the form of point clouds is advantageous and has inspired our work.

\subsubsection{Learning Without Direct Supervision.}
\sloppy In the context of autonomous driving, self-supervised learning has been used successfully for depth prediction~\cite{godard_digging_2019,yang_deep_2018}, as well as depth and ego-motion prediction~\cite{casser_depth_2019}.
Using only 2D supervision for 3D estimation is common in object reconstruction where the focus lies on estimating pose and shape for a diverse class of objects, but networks are commonly trained and evaluated on artificial datasets without noise. Generally, neural networks are trained to extract the 3D shape of an object from a single image. Initial works~\cite{yan_perspective_2016,kato_neural_2018} use multi-view images with known viewpoints to define a loss based on the ground truth segmentation mask in each image and the differentiably rendered shape. Subsequent methods~\cite{insafutdinov_unsupervised_2018,henderson_learning_2019} overcome the dependence on known poses by including the pose into the prediction pipeline and thus require only 2D supervision.

The aforementioned approaches rely on rendering a 2D image from the 3D representation to define loss functions based on the input. To enable training, the renderer has to be differentiable with respect to the 3D representation. Loper and Black~\cite{loper_opendr_2014} proposed a mesh-based, differentiable renderer called \opendr, which was extended in~\cite{henderson_learning_2019}. Other methods use approximations to ray casting for voxel volumes~\cite{yan_perspective_2016}, differentiable point clouds~\cite{insafutdinov_unsupervised_2018}, or differentiable rasterization for triangular meshes~\cite{kato_neural_2018}.

\subsection{Contribution}
We propose a monocular 3D vehicle detector that is trained without 3D bounding box labels by leveraging differentiable shape rendering. 
The major inputs for our model are 2D segmentation masks and depth maps, which we obtain from pre-trained, off-the-shelf networks. Therefore, our method does not require 3D bounding box labels for supervision. Two-dimensional ground truth and LiDAR point clouds are only required for training the pre-trained networks. We thus overcome the need for hand-labeled datasets which are cumbersome to obtain and contribute towards the wider applicability of 3D object detection. We train and evaluate the detector on the KITTI object detection dataset~\cite{geiger_are_2012}. The experiments show that our model achieves comparable results to state-of-the-art supervised monocular 3D detectors despite not using 3D bounding box labels for training. We further show that replacing the input monocular depth with stereo depth yields competitive stereo 3D detection performance, which shows the generality of our 3D detection framework.

\begin{figure}[tb]
  \centering
\includegraphics[width=\textwidth]{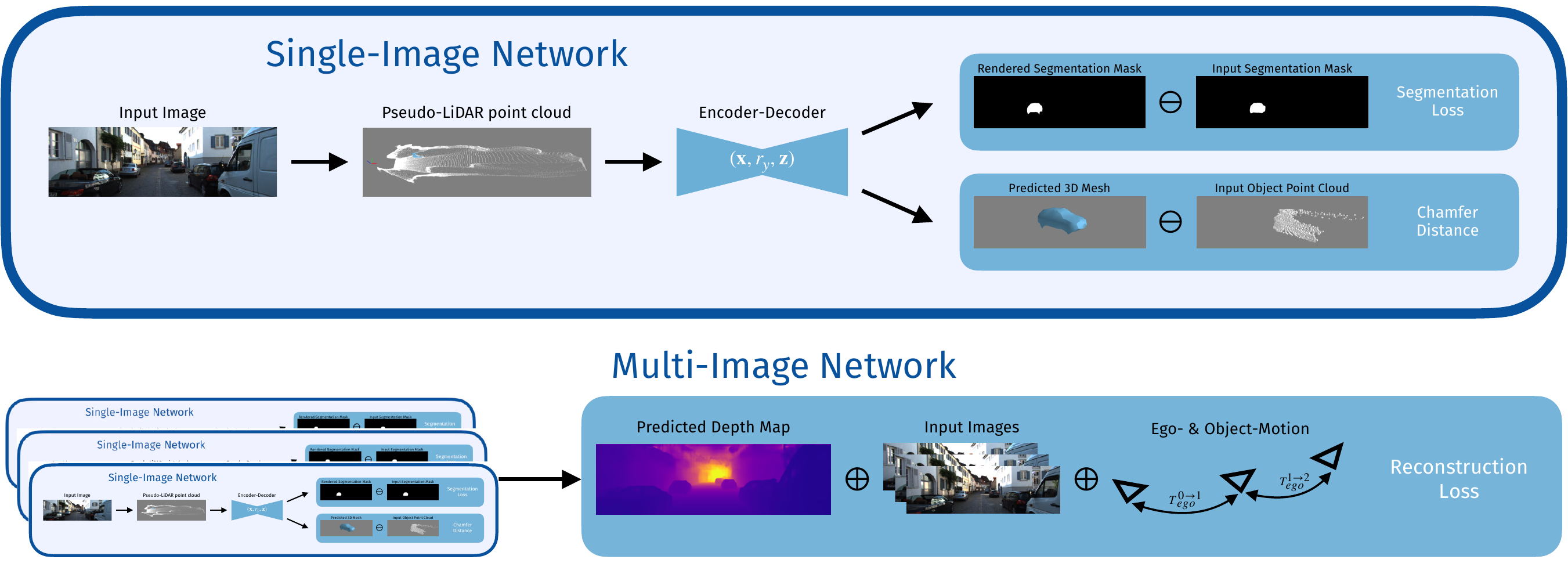}
\caption[Model architecture overview]{The proposed model contains a single-image network and a multi-image network extension. The single-image network back-projects the input depth map estimated from the image into a point cloud. A \fpointnet{} encoder predicts the pose and shape of the vehicle which are then decoded into a predicted 3D mesh and segmentation mask through differentiable rendering. The predictions are compared to the input segmentation mask and back-projected point cloud to define two loss terms. The multi-image network architecture takes three temporally consecutive images as the inputs, and the single-image network is applied individually to each image. Our network predicts a depth map for the middle frame based on the vehicle's pose and shape. A pre-trained network predicts ego-motion and object-motion from the images. The reconstruction loss is computed by differentiably warping the images into the middle frame.} \label{fig:model_model_architecture}
\end{figure}

\section{Learning 3D Vehicle Detection without 3D Bounding Box Labels} \label{sec:model}
The proposed model consists of a single-image network that can learn from single, monocular images and a multi-image extension that additionally learns from temporally consecutive frames. \autoref{fig:model_model_architecture} depicts the proposed architecture. We utilize pre-trained networks to compute depth maps, segmentation masks, and ego- and object-motion, which are used as inputs to the network and for the loss functions during training. During inference only the single-image network and the pre-trained image-to-depth and segmentation networks are required.

\subsection{Shape Representation} \label{subsec:model_shape_representation}
We use a mesh representation given by a mean mesh together with linear vertex displacements which are obtained from the manifold proposed in~\cite{engelmann_joint_2016} by a semi-manual process and are available on the project page. The mean vertex positions are denoted ${\vertexmean \in \R^{N \times 3}}$, the $K$ vertex displacement matrices are denoted ${\vertexbasis_k \in \R^{N \times 3}}, \; k = 1, \dots, K$, the shape coefficients are denoted ${\shape = \shapelong}$ and the deformed vertex positions in the canonical coordinate system are denoted ${\vertexdef \in \R^{N \times 3}}$. The deformed vertex positions are the linear combination
\begin{equation} \label{eqn:background_vertex_def}
    \vertexdef = \vertexmean + \sum_{k=1}^K z_k \cdot \vertexbasis_k \, .
\end{equation}

\begin{figure}[tb]
    \centering
    \begin{subfigure}[t]{0.2\textwidth}
        \centering
        \includegraphics[width=\textwidth]{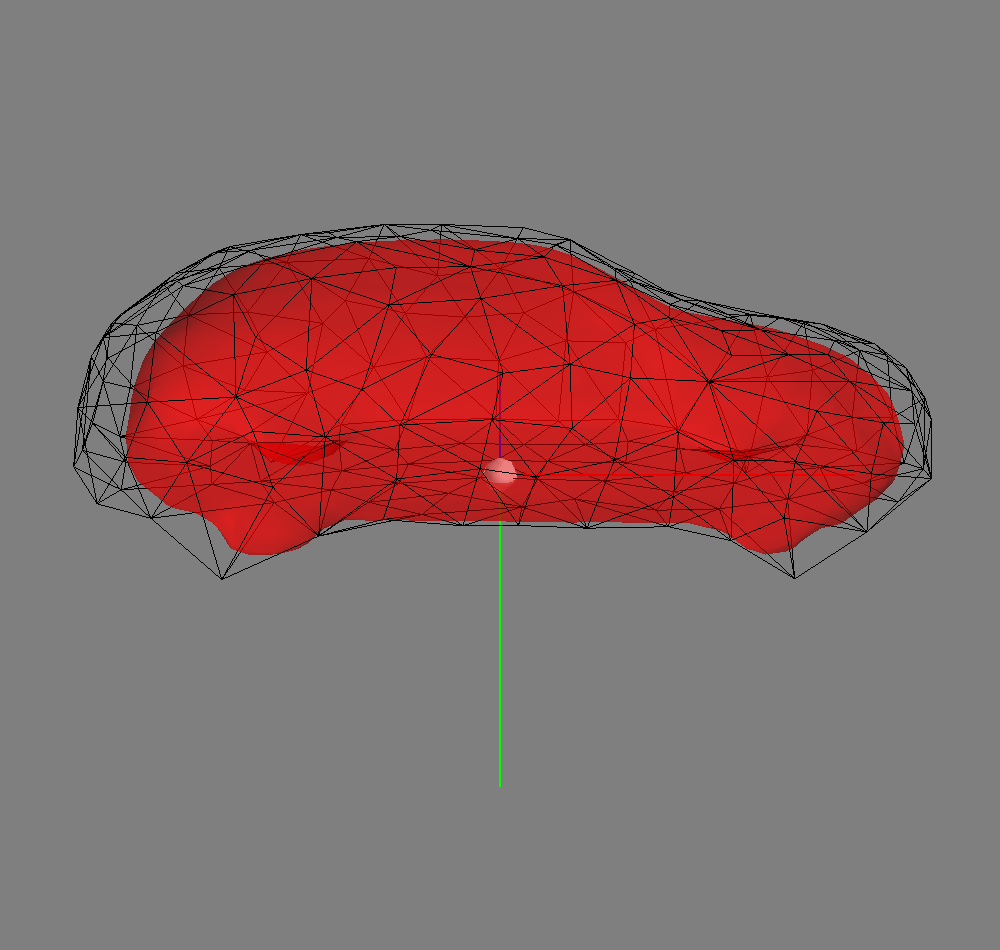}
        \caption{$\vertexmean + \vertexbasis_0$}
        \label{fig:model_mesh_basis0}
    \end{subfigure}
    \begin{subfigure}[t]{0.2\textwidth}
        \centering
        \includegraphics[width=\textwidth]{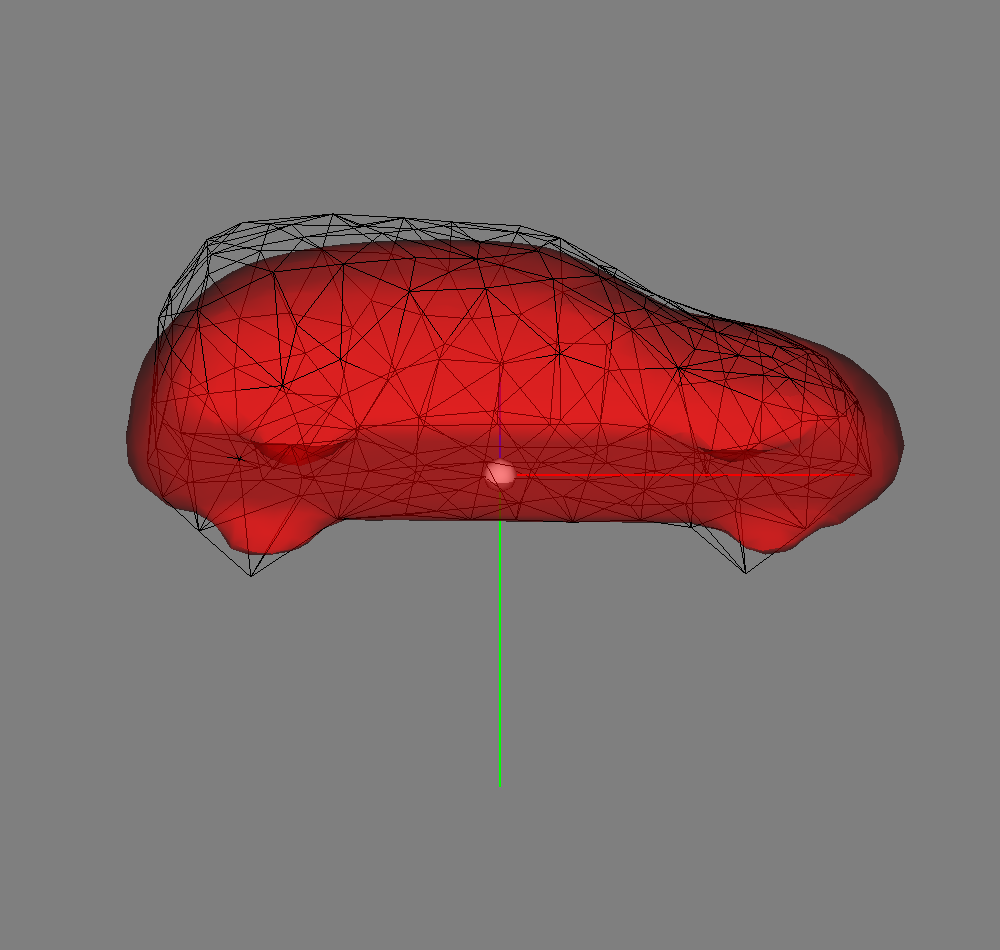}
        \caption{$\vertexmean + \vertexbasis_1$}
        \label{fig:model_mesh_basis1}
    \end{subfigure}
    \begin{subfigure}[t]{0.2\textwidth}
        \centering
        \includegraphics[width=\textwidth]{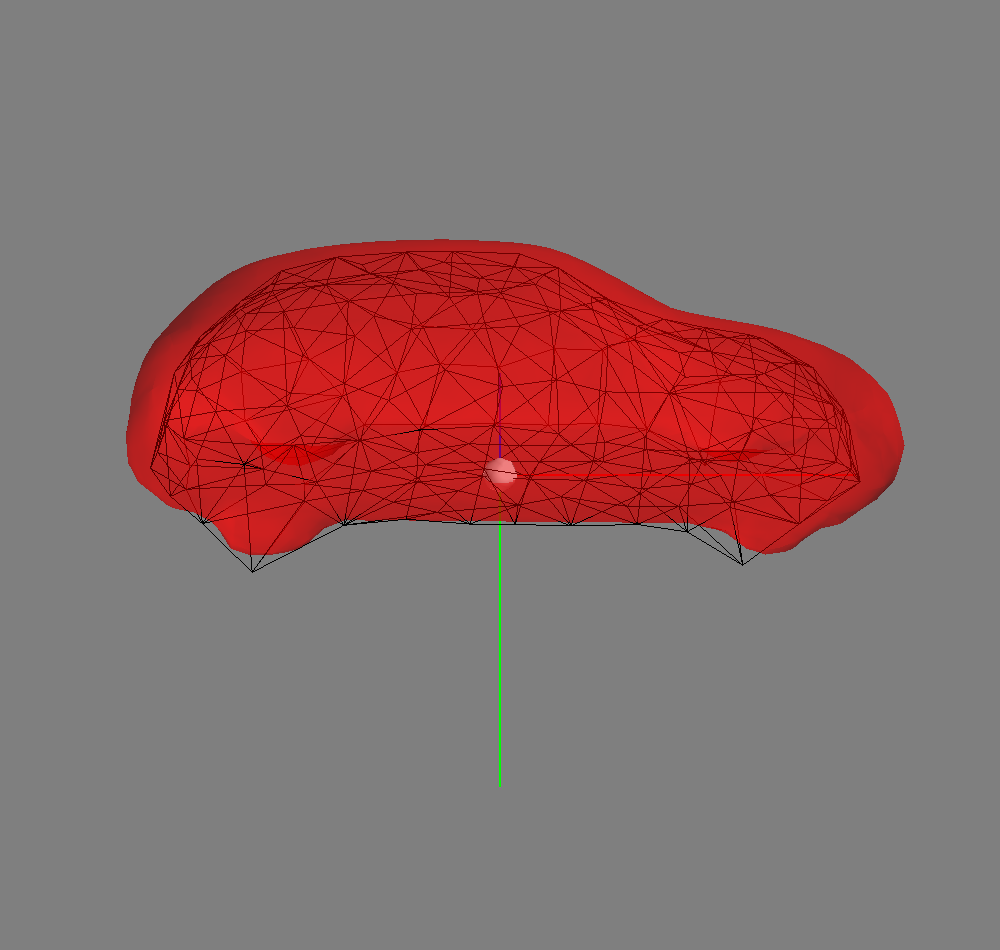}
        \caption{$\vertexmean + \vertexbasis_2$}
        \label{fig:model_mesh_basis2}
    \end{subfigure}
    \caption[Mesh mean shape and basis]{Shape manifold visualization. The mean shape is shown in red, and the deformed meshes are shown as black wireframes. The resulting shape space can represent longer (\ref{fig:model_mesh_basis0}), higher (\ref{fig:model_mesh_basis1}), and smaller (\ref{fig:model_mesh_basis2}) cars.}
    \label{fig:model_mesh_basis}
\end{figure}

\subsection{Single-Image Network}
The input depth map is back-projected into a point cloud, which decouples the architecture from the depth source as in \cite{wang_pseudo-lidar_2019}. The point cloud is filtered with the object segmentation mask to obtain the object point cloud. For depth maps from monocular images, the object point cloud frequently has outliers at occlusion boundaries, which are filtered out based on their depth values.

Afterward, a \fpointnet{} encoder~\cite{qi_frustum_2018} predicts the position $\position \in \R^3$, orientation $\ry \in [0, 2 \pi)$, and shape $\shape \in \R^K$ of the vehicle. The shape coefficients $\shape$ are applied in a canonical, object-attached coordinate system to obtain the deformed mesh based on our proposed shape manifold (\autoref{subsec:model_shape_representation}) using \autoref{eqn:background_vertex_def}. The deformed mesh is rotated by $\ry $ around the y-axis and translated by $\position$ to obtain the mesh in the reference coordinate system.

The deformed mesh in the reference coordinate system is rendered differentiably to obtain a predicted segmentation mask $\maskobject$ and a predicted depth map $\depthobject$. The rendered depth map $\depthobject$ that incorporates the predicted pose and shape of the vehicle is used only in the multi-image network. For the image areas which do not belong to the vehicle, as defined by the input segmentation mask, we utilize the input depth map as the background depth and render the depth from the deformed mesh otherwise. For rendering the predicted depth map and segmentation mask we utilize a recent implementation~\cite{henderson_learning_2019} of the differentiable renderer proposed in~\cite{loper_opendr_2014}. Additional details are in the supplementary material.

\subsection{Loss Functions}
\begin{figure}[tb]
\centering
\includegraphics[width=\textwidth]{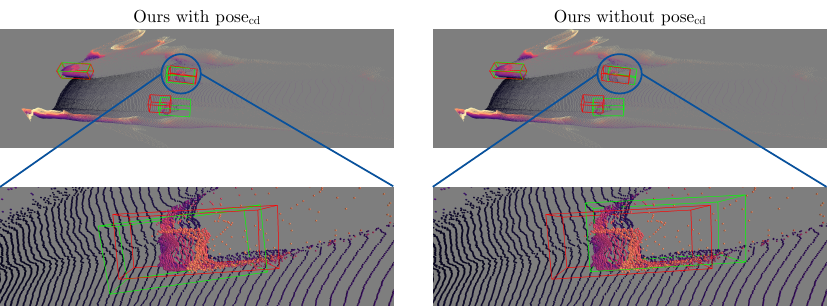}
\caption{Qualitative results with and without \posecd{} (cf. \autoref{subsec:loss_pose_cd}). We show the ground truth (\textit{red}) and the predictions (\textit{green}). Without the proposed \posecd{} the model learns to tightly fit the point cloud which leads to worse results due to errors in the point cloud. With \posecd{} the segmentation loss can correct the erroneous position of the point cloud and the predicted position is more accurate.}
\label{fig:qualitative_posecp}
\end{figure}

In order to train without 3D bounding box labels we use three losses, the segmentation loss $\lossseg$, the chamfer distance $\losscp$, and the photometric reconstruction loss $\lossreconstr$. The first two are defined for single images and the photometric reconstruction loss relies on temporal photo-consistency for three consecutive frames (\autoref{fig:model_model_architecture}). The total loss is the weighted sum of the single image loss for each frame and the reconstruction loss
\begin{equation} \label{eqn:loss_total}
    \losstotal  =  w_{rec} \cdot \lossreconstr + \frac{1}{3} \cdot \sum_t \losssingle^t \, ,
\end{equation}
where the single image loss is the weighted sum of the segmentation loss and chamfer distance
\begin{equation} \label{eqn:loss_single_image}
    \losssingle = 
    w_{cd}\cdot \losscp +
    w_{seg}\cdot \lossseg \, .
\end{equation}

To capture multi-scale information, the segmentation and reconstruction loss are computed for image pyramids~\cite{burt_laplacian_1983} with eight levels, which we form by repeatedly applying a $5 \times 5$ binomial kernel with stride two. For each pyramid level the loss values are the mean over the pixel-wise loss values which ensures equal weighting for each level.

\begin{figure}[tb]
    \centering
    \includegraphics[width=\textwidth]{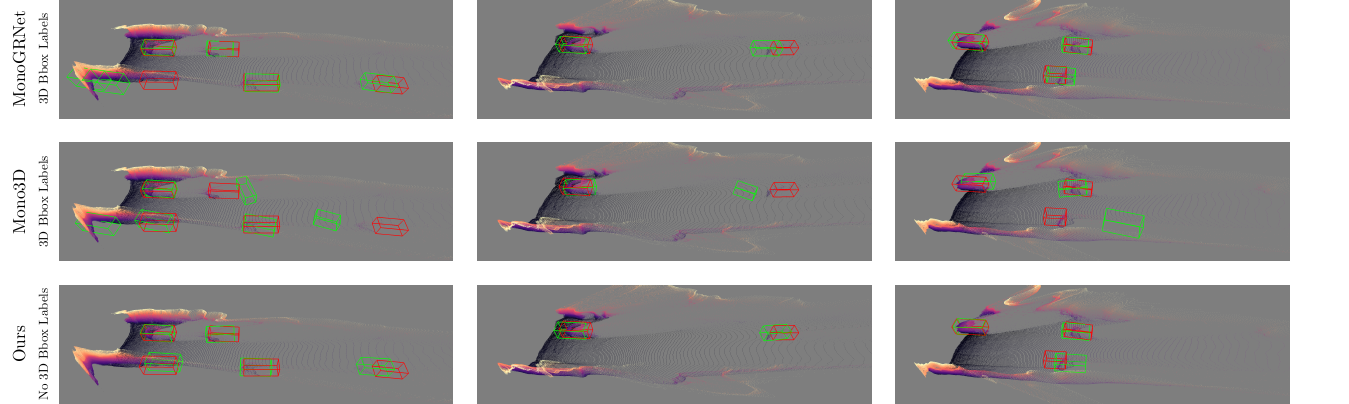}
    \caption[Qualitative comparison]{Qualitative comparison of \monogrnet~\cite{qin_monogrnet_2019} (\textit{first row}), \monothreed~\cite{chen_monocular_2016} (\textit{second row}), and our method (\textit{third row}) with depth maps from \bts~\cite{lee_big_2019}. We show ground truth bounding boxes for cars (\textit{red}), predicted bounding boxes (\textit{green}), and the back-projected point cloud. In comparison to \monothreed, the prediction accuracy of the proposed approach is increased specifically for further away vehicles. As in the quantitative evaluation (cf. \autoref{tab:comparison_state_of_the_art}), the performance of \monogrnet{} and our model is comparable.
    }
    \label{fig:qualitative_comparison_mono3d}
\end{figure}

\subsubsection{Segmentation Loss.} The segmentation loss penalizes the difference between the input segmentation mask $\maskobjectinput$  and the differentiably rendered segmentation mask $\maskobject$ using the squared $L^2$ norm.
\begin{equation}
    \lossseg = \fnorm{\maskobjectinput - \maskobject}^2 \, .
\end{equation}

\subsubsection{Chamfer Distance.}\label{subsec:loss_pose_cd}
The chamfer distance for point clouds, which was used in the context of machine learning by \cite{fan_point_2017}, penalizes the 3D distance between two point clouds. Its original formulation is symmetric w.r.t. the two point clouds. In contrast, the situation analyzed in this paper does not posses this symmetry. For each point $\vec{r}_i$ in the input object point cloud, there must exist a corresponding vertex $\vec{v}$ in the deformed mesh, while due to occlusion or truncation, the reverse is not true. Therefore, we use a non-symmetric version of the chamfer distance
\begin{equation}
    \losscp = \frac{1}{M} \sum_i \min_j \rho(\fnorm{\vec{r}_i - \vec{v}_{j}}) \, .
\end{equation}
We employ the Huber loss $\rho: \R \to \R^+_0$ to gain robustness against outliers.

For depth maps obtained from monocular image-to-depth networks, we notice weak performance of the chamfer distance (cf. \autoref{tab:ablation}) due to a high bias in the position of the input object point cloud, which is caused by the global scale ambiguity (cf. \autoref{fig:qualitative_posecp}). To use the orientation information captured in the object point cloud without deteriorating the position estimate, we introduce \posecd. The network outputs an auxiliary position $\vec{x}_{aux}$, and the chamfer distance is then calculated using this position
\begin{equation}
    \losscp = \losscp(\vec{x}_{aux}, \ry) \, .
\end{equation}
The auxiliary position $\vec{x}_{aux}$ is predicted by a separate network head. We cut the gradient flow between the main network and the additional head to not influence the main network, which necessitates the use of another loss term that back-propagates through the predicted position $\position$.

\subsubsection{Multi-Image Reconstruction Loss.}
The multi-image network is inspired by the recent success of self-supervised depth prediction from monocular images~\cite{casser_depth_2019,godard_digging_2019}, which relies on differentiably warping temporally consecutive images into a common frame to define the reconstruction loss. The single-image network is applied to three consecutive images $I^{t-1}, I^t, I^{t+1}$ of the same vehicle and the reconstruction loss is defined in the middle frame. The reconstruction loss is formulated as in \cite{casser_depth_2019} and we use their pre-trained network to estimate the ego-motion and object motion required for warping.

\subsubsection{Hindsight Loss.} \label{subsec:model_hindsight_loss}
To overcome the multi-modality of the loss w.r.t. the orientation of the vehicle, we apply the hindsight loss mechanism~\cite{guzman-rivera_multiple_2012}, which has been frequently used in the context of self-supervised object reconstruction~\cite{insafutdinov_unsupervised_2018,henderson_learning_2019}. The network predicts orientation hypotheses in $L$ bins and the hindsight loss is the minimum of the total loss over the hypotheses.

\setlength{\tabcolsep}{3.0pt}
\begin{table}[tb]
\begin{center}
\caption[KITTI evaluation]{Result for the proposed KITTI validation set. We report the average precision (AP) in percent for the car category in the bird's-eye view (BEV) and in 3D. The AP is the average over 40 values as introduced in~\cite{simonelli_disentangling_2019}. Our method convincingly outperforms the supervised baseline method \monothreed{} and shows promising performance in comparison to a state-of-the-art supervised method \monogrnet.}
\label{tab:comparison_state_of_the_art}
\begin{tabular}{l | c | c | c c c | c c c}
\toprule
\multicolumn{1}{c}{
\multirow{2}{*}{
\parbox[c]{.2\linewidth}{\centering Method}}} &
\multicolumn{1}{c}{\multirow{2}{*}{
\parbox[c]{.08\linewidth}{\centering Input}}} &
\multirow{2}{*}{
\parbox[c]{.11\linewidth}{\centering Without 3D Bbox}} &
\multicolumn{3}{c|}{AP\textsubscript{BEV, 0.7}} &
\multicolumn{3}{c}{AP\textsubscript{3D, 0.7}} \\
\cmidrule{4-6} \cmidrule{7-9}
\multicolumn{2}{c}{} &&   Easy  &  Mode  &  Hard  &  Easy  &  Mode  &  Hard   \\
\midrule
\midrule
Ours & Mono &  \checkmark        &   \zz{19.23} &    \zz{9.60} &    \zz{5.34} &    \zz{6.13} &    \zz{3.10} &    \zz{1.70} \\
\monogrnet~\cite{qin_monogrnet_2019} & Mono &     & \z 23.07 & \z 16.37 & \z 10.05 & \z 13.88 &  \z 9.01 &  \z 5.67 \\
\monothreed~\cite{chen_monocular_2016} & Mono &    &    1.92 &    1.13 &    0.77 &    0.40 &    0.21 &    0.17 \\
\bottomrule
\end{tabular}
\end{center}
\end{table}
\setlength{\tabcolsep}{1.4pt}

\section{Experiments}\label{sec:experiments}
We quantitatively compare our method with other state-of-the-art monocular 3D detection methods on the publicly available KITTI 3D object detection dataset~\cite{geiger_are_2012}. Note that since our method is the first monocular 3D detector trained without 3D bounding box labels, the compared-against methods are supervised methods that are trained with ground truth 3D bounding box labels. We conduct an extensive ablation study on the different loss terms to show the efficacy of each proposed component. Because the accuracy of the input point cloud plays a crucial role for the proposed model, we show experiments with depth maps estimated from different methods. Finally, we compare against methods based on per-instance optimization.

\begin{figure}[tb]
    \centering
    \includegraphics[width=\textwidth]{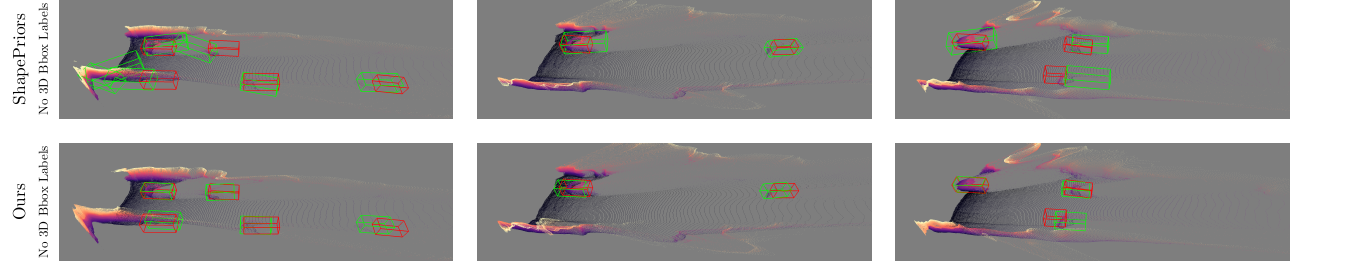}
    \caption[Qualitative comparison with ShapePriors]{Qualitative comparison of \shapepriors~\cite{engelmann_joint_2016} (first row) and our approach (second row) with depth maps from \bts. We show ground truth bounding boxes for cars (\textit{red}), predicted bounding boxes (\textit{green}), and the back-projected point cloud. \shapepriors{} is initialized with detections from \threedop~\cite{chen_3d_2015} as in the original paper, which leads to false positives (\textit{left column}). For the quantitative evaluation (cf. \autoref{subsec:experiments_shape_priors}) we control for this difference and our approach still shows better performance. The comparison shows that learning can produce more robust and accurate prediction than per-instance optimization. Both methods do not require 3D bounding box labels for training.
    }
    \label{fig:qualitative_comparison_shape_priors}
\end{figure}

\subsubsection{KITTI Object Detection.}
The KITTI dataset consists of sequences that are used for numerous benchmarks, e.g. 3D object detection and depth prediction. This leads to an overlap of the common validation set for object detection~\cite{chen_3d_2015} and the popular Eigen~\cite{eigen_split} train set for monocular depth estimation. The overlap was already noted by~\cite{wang_pseudo-lidar_2019}. Unlike in~\cite{wang_pseudo-lidar_2019}, we use a subset of the validation set that has no sequence-level overlap with the Eigen training set or the KITTI 2015 stereo training set. Following works can integrate pre-trained mono-to-depth and stereo-to-depth networks directly. The split files can be found on the project page. Results on the standard validation set~\cite{chen_3d_2015} are given in the supplementary material and they unsurprisingly show better performance than on the proposed split.

For the confidence score we estimate the KITTI category (\kittieasy, \kittimode, and \kittihard) from the data. We shift and scale the baseline scores $1 - \losssingle$ such that objects which are estimated to be \kittieasy{} have a higher score than any object which is estimated to be \kittimode. The same holds for \kittimode{} objects in comparison to \kittihard{} objects. This gives a slight improvement in average precision and details are in the supplementary material.

\setlength{\tabcolsep}{2.8pt}
\begin{table}[tb]
\begin{center}
\caption[Input depth]{Depth source ablation study. The average precision of the proposed model improves when using a supervised instead of an unsupervised image-to-depth method and when using stereo images instead of monocular images. Our more general method delivers the best performs among methods trained without 3D bounding box labels, but worse performance as the stereo-specific \stereorcnn{} which uses partial 3D bounding box information for training. Our approach clearly improves upon the common baseline \threedop{} and the recent \directshape{} and \tlnet.

\stereorcnn{} does not directly supervise the 3D position, but directly supervises the 3D bounding box dimensions. Additionally, they compute the viewpoint and perspective keypoint from the ground truth 3D bounding box label and use them for supervision and thus require 3D bounding box labels during training. Replacing the 3D bbox labels by estimated 3D dimensions, viewpoints, and perspective keypoints is a non-trivial extension of their work.}
\label{tab:comparison_depth}
\begin{tabular}{l | c | c | c c c | c c c}
\toprule
\multicolumn{1}{c}{
\multirow{2}{*}{
\parbox[c]{.2\linewidth}{\centering Method}}} &
\multicolumn{1}{c}{\multirow{2}{*}{
\parbox[c]{.08\linewidth}{\centering Input}}} &
\multirow{2}{*}{
\parbox[c]{.11\linewidth}{\centering Without 3D Bbox}} &
\multicolumn{3}{c|}{AP\textsubscript{BEV, 0.7}} &
\multicolumn{3}{c}{AP\textsubscript{3D, 0.7}} \\
\cmidrule{4-6} \cmidrule{7-9}
\multicolumn{2}{c}{} &&   Easy  &  Mode  &  Hard  &  Easy  &  Mode  &  Hard   \\
\midrule
\midrule
Ours (Monodepth) & Mono & \checkmark  &   10.78 &    5.43 &    2.99 &    4.53 &    2.16 &    1.17 \\
Ours (\bts{}) & Mono   & \checkmark      &   19.23 &    9.60 &    5.34 &    6.13 &    3.10 &    1.70 \\
\midrule
Ours (\sgm{}) & Stereo & \checkmark    &   31.51 &   15.78 &    8.76 &    8.42 &    4.08 &    2.23 \\
Ours (\ganet{}) & Stereo & \checkmark &  \zz{68.16} &  \zz{35.82} &  \zz{20.45} &  \zz{38.45} &  \zz{18.78} &  \zz{10.44} \\
 \midrule
\stereorcnn~\cite{li_stereo_2019} & Stereo & (\checkmark)  & \z 71.51 &  \z 53.81 & \z 35.56 & \z 56.68 & \z 38.30 & \z 25.45 \\
\tlnet~\cite{qin_triangulation_2019} & Stereo &       &   24.92 &   17.01 &   11.25 &   13.74 &    9.45 &    6.13 \\
\directshape~\cite{wang2020directshape} & Stereo & \checkmark   &   24.91 &   16.03 &   10.28 &   12.60 &    7.36 &    4.33 \\
\threedop~\cite{chen_3d_2015} & Stereo &    &    8.72 &    5.52 &    3.29 &    2.68 &    1.48 &    1.05 \\
\bottomrule
\end{tabular}
\end{center}
\end{table}
\setlength{\tabcolsep}{1.4pt}

\subsubsection{Pre-Trained Networks.}
For \maskrcnn~\cite{he_mask_2017} we use the implementation of \cite{matterport_maskrcnn_2017} and their pre-trained weights on the COCO \cite{coco} dataset.
For ego- and object-motion estimation we utilize the official implementation of \structtodepth~\cite{casser_depth_2019} and their pre-trained weights on the Eigen train split.
For depth estimation we use \monodepth~\cite{godard_digging_2019}, \bts~\cite{lee_big_2019}, \sgm~\cite{hirschmuller_accurate_2005}, and \ganet~\cite{zhang_ga-net:_2019}.
For \monodepth{} we use the official implementation and their pre-trained weights on Zhou's~\cite{sfm_learner} subset of the Eigen train split; this model is trained with supervision from monocular images of resolution $1024\times320$ and utilizes pre-training on ImageNet~\cite{imagenet}.
For \bts{} we use the official implementation and their pre-trained weights on the Eigen train split.
For \sgm{} we use the public implementation provided by \cite{libSGM2020} and piecewise linear interpolation in 2D to complete the disparity map.
For \ganet{} we use the official implementation and their pre-trained weights on Scene~Flow~\cite{scene_flow} and the KITTI 2015 stereo training set.
For matching consecutive segmentation masks we use a similar procedure to~\cite{casser_depth_2019}; however, we first warp the segmentation masks into a common frame using optical flow~\cite{yin_hierarchical_2019}.

\subsubsection{Evaluation Results.}
For monocular object detection, we compare to two supervised monocular 3D detection networks: \monogrnet~\cite{qin_monogrnet_2019} is a state-of-the-art monocular detector and \monothreed~\cite{chen_monocular_2016} is a common baseline method. \autoref{tab:comparison_state_of_the_art} shows the evaluation results. Our results are superior to the ones generated by \monothreed{} in all categories. While \monogrnet{} outperforms our method, the performance gap is relatively small. This difference is smaller for the \kittieasy{} category than for the \kittimode{} category, which shows that handling distant objects and occlusions when learning without 3d bounding box labels is challenging.

\setlength{\tabcolsep}{4pt}
\begin{table}[tb]
\begin{center}
\caption[Ablation Study]{Ablation study using depth maps from \bts~\cite{lee_big_2019}. Using the chamfer distance without the proposed \posecd{} reduces the accuracy significantly. Learning pose and shape without 3D bounding box labels is an under-constraint problem and the performance decreases (cf. last row). Without multi-image training the performance in the BEV is similar but the performance in 3D is decreased.}
\label{tab:ablation}
\begin{tabular}{l | c c c | c c c}
\toprule
\multirow{2}{*}{
\parbox[c]{.2\linewidth}{\centering Method}}
& \multicolumn{3}{c|}{AP\textsubscript{BEV, 0.7}} &
  \multicolumn{3}{c}{AP\textsubscript{3D, 0.7}} \\ 
  \cmidrule{2-4} \cmidrule{5-7}
&  Easy  &  Mode  &  Hard  &  Easy  &  Mode  &  Hard   \\
\midrule
\midrule
Full Model                  & \zz{19.23} &  \z 9.60 &  \z 5.34 &  \z 6.13 &  \z 3.10 &  \z 1.70 \\
W/o $\losscp$               & 9.75 &    5.21 &    2.75 &    3.50 &    1.73 &    0.98 \\
W/o \posecd                 & 4.53 &    2.84 &    1.58 &    0.94 &    0.48 &    0.26 \\
W/o $\lossseg$              & 4.22 &    2.23 &    1.16 &    0.76 &    0.41 &    0.18 \\
W/o $\lossreconstr$             & \z19.60 &    \zz{9.48} &    \zz{5.30} &    4.88 &    2.26 &    1.20 \\
W/\phantom{o} $B_k$      &   16.02 &    8.12 &    4.51 &    \zz{5.24} &    \zz{2.59} &    \zz{1.32}\\
  \bottomrule
  \end{tabular}
\end{center}
\end{table}
\setlength{\tabcolsep}{1.4pt}

\subsection{Ablation Study}
\subsubsection{Input Depth.}
\autoref{tab:comparison_depth} shows that the average precision with \bts~\cite{lee_big_2019}, a supervised mono-to-depth network, is better than the performance with the self-supervised \monodepth~\cite{godard_digging_2019}, due to the superior depth estimation accuracy.
This leads to the question: \textit{Does the performance of the proposed model constantly improve if more accurate depth maps are used as input?}
When switching from mono to stereo, better depth maps are estimated, and the AP is dramatically improved, as can be seen in \autoref{tab:comparison_depth}.
Besides, using depth maps from \ganet~\cite{zhang_ga-net:_2019}, a stereo-to-depth network trained in a supervised fashion, outperforms using depth maps from the traditional stereo matching algorithm \sgm~\cite{hirschmuller_accurate_2005} by a notable margin.
In~\autoref{tab:comparison_depth}, we also show the results of state-of-the-art stereo 3D detectors, \stereorcnn~\cite{li_stereo_2019}, \directshape~\cite{wang2020directshape}, \threedop~\cite{chen_3d_2015}, and \tlnet~\cite{qin_triangulation_2019}. The proposed approach ranks first among the methods that do not use 3D bounding box labels for training.

\subsubsection{Loss Terms.}
We demonstrate the significance of using the chamfer distance together with the proposed \posecd{} in \autoref{fig:qualitative_posecp} and \autoref{tab:ablation}. Simultaneously estimating pose and shape generally resulted in worse performance and training instabilities due to the inherent scale ambiguity. The best results we achieved are obtained with the mean shape -- the shape variability of cars within the KITTI dataset is small and thus a fixed shape is a reasonable approximation. More details can be found in the supplementary material. During our experiments, the reconstruction loss in the multi-image setting contributes marginal improvements, which may be due to the noise in the ego-motion and object-motion predictions, which were taken from the self-supervised \structtodepth~\cite{casser_depth_2019}; details are included in the supplementary material.

\subsection{Comparison with Non-Learning-Based Methods}
\label{subsec:experiments_shape_priors}
We choose \shapepriors~\cite{engelmann_joint_2016} for comparison because it uses very similar input data; \shapepriors{} uses depth maps and initial 3D detections, while our method uses depth maps and 2D segmentation masks during inference. We compare both methods using depth maps generated by \ganet.

The initial 3D detections were taken from \threedop{} in the original paper. To facilitate a fair quantitative comparison, we initialize the position with the median of the object point cloud in the $x$ and $z$ direction and the minimum in the $y$ direction. For the orientation and the 2D bounding box we use the ground truth. Because we require the ground truth label for the orientation initialization and the segmentation mask for the position initialization, we match segmentation masks and labels. Thus, the results presented here are not comparable to the other results within this paper.

Under these conditions, \shapepriors{} achieves $23.65\%$ AP\textsubscript{BEV, 0.7, easy} and ours $77.47 \%$. For the qualitative comparison (cf. \autoref{fig:qualitative_comparison_shape_priors}) \shapepriors{} is initialized with detections from \threedop~\cite{chen_3d_2015} as in the original paper. The quantitative and qualitative comparisons show that per-instance optimization delivers less robust and accurate predictions than learning.
Similarly, the comparison against \directshape{} (cf. \autoref{tab:comparison_depth}) indicates that learning can extract meaningful priors from the training data and ultimately deliver superior performance.

\section{Conclusion}
We propose the first monocular 3D vehicle detection method for real-world data that can be trained without 3D bounding box labels. By proposing a differentiable-rendering-based architecture we can train our model from unlabeled data using pre-trained networks for instance segmentation, depth estimation, and motion prediction. During inference only the instance segmentation and depth estimation networks are required. Without ground truth labels for training, we decisively outperform a baseline supervised monocular detector and show promising performance in comparison to a state-of-the-art supervised method.

Furthermore, we demonstrate the generality of the proposed framework by using depth maps from a stereo-to-depth network and without further changes achieving state-of-the-art performance for stereo 3D object detection without 3D bounding box labels for training.
While this paper demonstrates that monocular 3D object detection without 3D bounding box labels for training is viable, many directions for future research remain, e.g. the explicit integration of stereo images, the extension to pedestrians and cyclists, training on large, unlabelled datasets, or the integration of an occlusion aware segmentation loss.

\clearpage
\section*{Acknowledgement}
We thank the anonymous reviewers for providing helpful comments. We express our appreciation to Artisense for supporting this work. This work has been presented at the German Conference for Pattern Recognition (GCPR) 2020 and the final paper will be published in LNCS (Springer). The final authenticated publication will be referenced upon publication.

\bibliographystyle{splncs03}
\bibliography{main}

\end{document}

%% file: commands.tex
\newcommand{\br}[1]{\left(#1\right)}

\newcommand{\R}{\mathbb{R}}

\newcommand{\be}{\begin{equation*}}
\newcommand{\ee}{\end{equation*}}

\def\ba#1\ea{\begin{align*}#1\end{align*}}

\newcommand{\fnorm}[1]{\left|\left| #1 \right| \right|}

\newcommand{\mksym}[3]{%
\expandafter\newcommand\csname #1\endcsname{#2}%
}

\newcommand{\vc}[1]{\boldsymbol{#1}}
\renewcommand{\vec}[1]{\vc{#1}}

\mksym{position}{\ensuremath{\vec{x}}}{%
Position $\br{x, y, z}$ of the object}

\mksym{ry}{\ensuremath{r_y}}{%
Orientation around $y$-axis of the object}

\mksym{roty}{\ensuremath{\vec{R}_y}}{%
Rotation matrix corresponding to $\ry$}

\mksym{pose}{\ensuremath{\vec{p}}}{%
Pose $\br{x, y, z, \ry}$ of the object}

\mksym{shape}{\ensuremath{\vc{z}}}{%
Shape coefficients $\br{z_1, \dots, z_K}$ of the object}

\newcommand{\shapelong}{\br{z_1, \dots, z_K}}

\mksym{image}{\ensuremath{I}}{%
Image}

\mksym{depthinput}{\ensuremath{D_{in}}}{%
Input depth map}

\mksym{pcinput}{\ensuremath{\vec{P}}}{%
Point cloud of the input depth map}

\mksym{pcobject}{\ensuremath{\vec{P}_{obj}}}{%
Point cloud of the object}

\mksym{maskobjectinput}{\ensuremath{S_{in}}}{%
Input object segmentation mask}

\mksym{egomotion}{\ensuremath{T_{ego}}}{%
Ego-motion}

\mksym{objectmotion}{\ensuremath{T_{obj}}}{%
Object motion}

\mksym{depthobject}{\ensuremath{D_{obj}}}{%
Predicted depth map}

\mksym{maskobject}{\ensuremath{S_{obj}}}{%
Predicted object segmentation mask}

\mksym{vertexmean}{\ensuremath{\vec{V}_{0}}}{%
Mean mesh vertex positions (row-wise)}

\mksym{vertexbasis}{\ensuremath{\vec{B}}}{%
Mesh vertex displacement basis matrices (row-wise)}

\mksym{vertexdef}{\ensuremath{\vec{V}}_{def}}{%
Deformed mesh vertex positions (row-wise)}

\mksym{vertexnormalsmean}{\ensuremath{\vec{N}_{0}}}{%
Mean mesh vertex normals (row-wise)}

\mksym{vertexnormalsbasis}{\ensuremath{\vec{dN}}}{%
Mesh vertex normals basis matrices (row-wise)}

\mksym{vertexnormalsdef}{\ensuremath{\vec{N}_{def}}}{%
Deformed mesh vertex normals (row-wise)}

\mksym{vertexnormalsdefworld}{\ensuremath{\vec{N}_{def, cam}}}{%
Deformed mesh vertex normals in world coordinates (row-wise)}

\mksym{losstotal}{\ensuremath{\mathcal{L}_{tot}}}{%
Total Loss}

\mksym{losshindsight}{\ensuremath{\mathcal{L}_{hs}}}{%
Hindsight Loss}

\mksym{losscp}{\ensuremath{\mathcal{L}_{cd}}}{%
Chamfer distance/loss}

\mksym{lossseg}{\ensuremath{\mathcal{L}_{seg}}}{%
Segmentation Loss}

\mksym{lossreconstr}{\ensuremath{\mathcal{L}_{rec}}}{%
Reconstruction Loss}

\mksym{losssingle}{\ensuremath{\mathcal{L}_{single}}}{%
Single image loss}

\mksym{averageprecision}{\ensuremath{AP}}{%
Average precision}

\newcommand{\mkgls}[3]{%
\expandafter\newcommand\csname #1\endcsname{#2}%
}

\mkgls{tnet}{T-Net}{%
Part of Frustum PointNet proposed by \cite{qi_frustum_2018}}

\mkgls{boxnet}{Box-Net}{%
Our box estimation network shown in \autoref{fig:model_network_architecture}}

\mkgls{posecd}{pose\textsubscript{cd}}{%
Modification to the ICP loss that uses a decoupled position for the ICP loss to deal with bias in the point cloud (see \autoref{sec:loss_pose_icp})}

\mkgls{fasterrcnn}{Faster R-CNN}{%
Faster R-CNN proposed by \cite{ren_faster_2015}}

\mkgls{maskrcnn}{Mask R-CNN}{%
Mask R-CNN proposed by \cite{he_mask_2017}}

\mkgls{adam}{ADAM}{%
ADAM optimizer proposed by \cite{kingma_adam:_2015}}
\mkgls{pointnet}{PointNet}{%
PointNet proposed by \cite{qi_pointnet:_2017}}

\mkgls{fpointnet}{Frustum PointNet}{%
Frustum PointNet proposed by \cite{qi_frustum_2018}}

\mkgls{pseudolidar}{Pseudo-LiDAR}{%
Pseudo-LiDAR proposed by \cite{wang_pseudo-lidar_2019}}

\mkgls{pseudolidarpp}{Pseudo-LiDAR++}{%
Pseudo-LiDAR++ proposed by \cite{you_pseudo-lidar++:_2019}}

\mkgls{threedop}{3DOP}{%
3D Object Proposals for Accurate Object Class Detection \parencite{chen_3d_2015}}

\mkgls{shapepriors}{ShapePriors}{%
Joint Object Pose Estimation and Shape Reconstruction in Urban Street Scenes Using 3D Shape Priors \parencite{engelmann_joint_2016}}

\mkgls{avod}{AVOD}{%
Aggregate View Object Detection proposed by \cite{ku_joint_2018}}

\mkgls{stereorcnn}{Stereo-RCNN}{%
Stereo-RCNN proposed by \cite{li_stereo_2019}}

\mkgls{directshape}{DirectShape}{%
DirectShape proposed by \cite{wang_directshape:_2019}}

\mkgls{tlnet}{TLNet}{%
Triangulation Learning Network: from Monocular to Stereo 3D Object Detection \parencite{qin_triangulation_2019}}

\mkgls{monothreed}{Mono3D}{%
Mono3D proposed by \cite{chen_monocular_2016}}

\mkgls{deepthreedbox}{Deep3DBox}{%
Deep3DBox proposed by \cite{mousavian_3d_2017}}

\mkgls{deepmanta}{Deep MANTA}{%
Deep MANTA proposed by \cite{chabot_deep_2017}}

\mkgls{multifusion}{MLF}{%
Multi-Level Fusion based 3D Object Detection from Monocular Images \parencite{xu_multi-level_2018}}

\mkgls{threedrcnn}{3D-RCNN}{%
3D-RCNN proposed by \cite{kundu_3d-rcnn:_2018}}

\mkgls{monogrnet}{MonoGRNet}{%
Monocular geometric reasoning network proposed by \cite{qin_monogrnet:_2019}}

\mkgls{lasernet}{LaserNet}{%
LaserNet proposed by \cite{meyer_lasernet:_2019}}

\mkgls{ganet}{GA-Net}{%
Guided Aggregation Net proposed by \cite{zhang_ga-net:_2019}}

\mkgls{sgm}{SGM}{%
Semi Global Matching proposed by \cite{hirschmuller_accurate_2005}}

\mkgls{bts}{BTS}{%
From Big to Small: Multi-Scale Local Planar Guidance for Monocular Depth Estimation \parencite{lee_big_2019}}

\mkgls{monodepthone}{Monodepth}{%
Monodepth proposed by \cite{godard_unsupervised_2017}}

\mkgls{monodepth}{Monodepth 2}{%
Monodepth 2 proposed by \cite{godard_digging_2019}}

\mkgls{structtodepth}{struct2depth}{%
Depth Prediction Without the Sensors: Leveraging Structure for Unsupervised Learning from Monocular Videos \parencite{casser_depth_2019}}

\mkgls{hdthree}{HD\textsuperscript{3}}{%
Hierarchical Discrete Distribution Decomposition for Match Density Estimation \parencite{yin_hierarchical_2019}}

\mkgls{shapenet}{ShapeNet}{%
ShapeNet proposed by \cite{chang_shapenet:_2015}}
\mkgls{kittiobject}{KITTI Object}{%
KITTI object detection benchmark proposed by \cite{geiger_are_2012}}

\mkgls{kittieasy}{easy}{%
Category "easy" for the \kittiobject{} benchmark (see \autoref{sec:background_kitti_categories})}

\mkgls{kittimode}{moderate}{%
Category "moderate" for the \kittiobject{} benchmark (see \autoref{sec:background_kitti_categories})}

\mkgls{kittihard}{hard}{%
Category "hard" for the \kittiobject{} benchmark (see \autoref{sec:background_kitti_categories})}

\mkgls{valclean}{val clean}{%
Validation set proposed within this thesis}

\mkgls{opendr}{OpenDR}{%
Open differentiable rendering proposed by \cite{loper_opendr:_2014}}

\mkgls{dirt}{DIRT}{%
DIRT: a fast differentiable renderer for TensorFlow \parencite{dirt2020}}

\mkgls{tensorflow}{TensorFlow}{%
Deep learning framework proposed by \cite{tensorflow2015-whitepaper}}

\newcommand*{\fullref}[1]{\hyperref[{#1}]{\autoref*{#1} (\nameref*{#1})}}

\newcommand\z{\bfseries}

\newcommand{\zz}[1]{\underline{#1}}